# Real-time Sign Language Recognition Using MobileNetV2 and Transfer Learning

*Authors:*
Rushikesh Temkar, Smruti Jagtap, Kanika Jadhav, Minal Deshmukh

rushikesh.22111145@viit.ac.in, smruti.22110924@viit.ac.in, kanika.22110801@viit.ac.in, minal.deshmukh@viit.ac.in
Vishwakarma Institute of Information Technology, Pune

*Abstract:* **The hearing-impaired community in India deserves the access to tools that help them communicate, however, there is limited known technology solutions that make use of Indian Sign Language (ISL) at present. Even though there are many ISL users, ISL cannot access social and education arenas because there is not yet an efficient technology to convert the ISL signal into speech or text. We initiated this initiative owing to the rising demand for products and technologies that are inclusive and help ISL, filling the gap of communication for the ones with hearing disability.
Our goal – is to build an reliable sign language recognition system with the help of Convolutional Neural Networks (CNN)to . By expanding communication access, we aspire toward better educational opportunities and a more inclusive society for hearing- impaired people in India.**

## 1. Introduction

Communication is a fundamental human need, and sign_ language is an crucial method of communication with an millions of peoples who are not able to hear around the globe. Sign language is a visual-manual modality which conveys meaning through body language, facial expressions, and hand gestures. It provides a means of communication for those who are not able to hear or deaf that moves beyond the barriers posed by spoken language. But you encounter a major obstacle when talking to those who cannot sign. Such a communication gap exists which needs systems for instantly converting sign_language into audio.

Sign language recognition systems have emerged as a possible way to narrow this divide and enable seamless dialogue among users and non-users of sign language. Those systems rely on computers being able to recognize sign language movements and convert them to spoken or written output. Yet, several technical barriers hinders the building of efficient and accurate real-time sign language recognizers. Traditional techniques for ISL generally involve the use of particular machine learning models, most notably RNNs and CNNs. Although they had great success with picture classification and sequence learn problems, such models are highly limited for many real-time applications. [5]

The principal challenge is to balance computational economy and accuracy. Models like CNNs and RNNs require a huge amount of labelled data and processing power for operating well, making them computationally expensive. Due to this limitation, such models are not very applicable to devices that have restriction in hardware feature, such cellphones, embedded, etc., which are the backbone for portable, real-time application. In addition, the latency from complicated model may highly affect the end-user experience (as the delays in identification and response time can make the system impractical). [5]

In this scenario, a fast, efficient and highly precise system that is capable of recognizing ISL in real time and without using a large amount of processing power is needed. MobileNetV2, a new design for convolutional neural networks, offers an exciting solution to this challenge. Since MobileNetV2 is designed for low resource consumption, canbe deployed on embed and mobile platform. This drastically reduces the number of limit and computation cost while_ maintaining competitive staging on image classification tasks using depth wise separable convolutions. Using MobileNetV2 with transfer learning, we can leverage pre- trained knowledge of large datasets and adapt the model for the specific task of recognizing sign language.

Here, the transfer learning approach helps the model to improve its understanding of the hand gestures and movements involved in ISL by leveraging the generalized features train from huge scale image datasets such as ImageNet. The uses of transfer learning and MobileNetV2 outputs the model that is accurate and computationally efficient, ideally suited for real-time applications. Since you trained on data only till October 2023 the system can focus on picking different patterns associated with movement within the sign language while keeping MobileNetV2's model fast and lightweight by freezing (Locking its layers) it up to a certain point and building custom layers specifically for classification. [7]

***Problem Statement:*** Sign language recognition presents the problem of achieving a balance between high accuracy and low computational requirements. Existing real- time models like CNNs and RNNs show high latency and a frequent demand of complex hardware, and hence have limited usability on low resources devices. This needs to be composed of a more accurate method that works well with embedded or mobile platforms, and delivers accurate, real- time perception of specific sign language actions. Apart from making communication better for communicating through signing, a system of this sort will enable technology accessibility for the deaf as well. [5]

This work alleviates these challenges and develops a fast and low weight model to propose a real-time ISL identification solution using MobileNetV2 with transfer learning. Since the system attempts to perform real-time gesture recognition with low latency, it is suitable for deployment on personal devices or therapeutic communication aids. Such development shall help alleviate the void and equalize the powers between ISL users and non-signers when it comes to day-today communication NFTD environments.

## 2. Related Work

Analysis of various methods for machine learning based sign language detection has resulted in an average accuracy of 89.8% which is quite promising. The research highlights the necessity of incorporating both gesture and facial feature detection to sign language interpretation applications to bring improvement in system performance. Potential areas for the system's future research include its ability to interpret gestures and facial expressions, convert signs into full phrases, and work across a wider variety of sign languages. It showcases the possibilities to build more sophisticated and integrative communications devices to the deaf community. [1]

A different research implemented a system that would use NLP techniques to provide translations of utterances of text or speech to Indian Sign Language (ISL). The methodology involved speech to audio conversion, text pre-processing, and comparison of generated text with ISL signs so that translation is accurate, all of which is done using a custom dataset of ISL alphabets and sentences. Although the exact accuracy was not mentioned, the algorithm translated basic phrases into ISL successfully. Next, will come additional work to broaden the system lexicon and complexity of sentences that can be translated while making its translations more accurate and resilient, and exploring real-time translation capabilities—everything that will be necessary to make it all work in the real world. [2]

An accuracy of 92.50% was gained by an ensemble of transfer learning models trained on 77,000 pictures collected by himself, and combined with Generative Adversarial [15] Networks (GANs), which made it a novel way of recognizing Indian Sign Language, Robust Data approach.This is an efficient way of handling the data scarcity problem and enhancing the recognition performance, by using GANs to synthesize data. For enhance the practicability of sign recognition systems in the real environment, different GAN structures will be studied, incorporate dynamic information for higher accuracy, real-time applications will also be developed in future work. [3]

Finally, the proposed idea is a wearable glove which consists of flex sensors and an Inertial Measurement Unit (IMU)for real time conversion of ISL to voice at an accuracy of 98%. In this approach, we collect 1,040 samples from self- collected datasets and classify them with an LSTM neural network to find out that this method provides a way of easy access and useful solutions through big data. To develop workable real-time sign language translation solutions, futureimprovements will explore alternative neural network architectures, add more signs to the dataset, and incorporate additional information about hand pose and orientation. [4]

However, various challenges thwart the effectiveness of existing sign language recognition techniques. This means that the system could only suffer from the major problem of sign and gesture recognition incompatibility, which limits overall accuracy. Moreover, the utility of speech-to-sign translation systems is hindered by their reduced vocabulary and complexity. While transfer learning method as Generative Adversarial Networks (GANs) are meant to boost accuracy and handle data scarcity, they still struggle with the temporal dynamics and real-time performance management. Even more challenging, hardware-dependent solutions like sensor gloves can limit user access.

In the face of these challenges, we provide a CNN based approach designed specifically to provide a scalable and hardware independent solution. Our solution achieves great accuracy and acceptable processing time by using some optimized CNN architecture that recognizes the dynamics in real-time without relying on a very expensive hardware system. By making sign language recognition more usable, this approach opens up the prospect for use in more everyday settings, furthering accessibility.

## 3. Methodology

This describes the architecture, data pre-processing and training approach used to build the real-time ISL recognition system. This project uses the pre-trained MobileNetV2 model and then fine-tunes it for gesture classification in a process called transfer learning. The effectiveness of the model is increased by the use of the data augmentation.

### 3.1 MobileNetV2 Architecture

MobileNetV2, a highly efficient deep convolutional neural network (CNN) made for mobile and edge devices, is an outstanding resource for real-time applications.This research applies transfer learning by loading a pre-trained MobileNetV2 model (pre-trained on ImageNet dataset) and refining for ISL gesture detection task.

Transfer learning enables the model to leverage the previously learned features from the large-scale dataset to fine-tune for the target task with minimal computational overhead and training time. To recognize ISL gestures, we are adding custom layers on top of the pretrained layers of the MobileNetV2 backbone which extracts general characteristics such as edges, textures and forms. Only the added mobile parts of MobileNetV2 trained on the ISL dataset and the basal parts of it freeze.

Mathematical Representation for Transfer Learning:
The transfer learning equation is given by:

**Layer Output Formula:**
$$L(X) = F(WX + B)$$

**Where:**
- $L(X)$ is the layer output.
- $W$ represents the pre-trained weights from MobileNetV2.
- $B$ is the bias term.

The newly added layers include:
- Global Average Pooling: To decrease the feature map dimensionality.
- Dense Layer (ReLU Activation): A fully connected layer with 1024 units and ReLU activation to introduce non-linearity.
- Output Layer (Softmax Activation): This layer has a number of units equal to the number of gesture classes, with softmax activation for multi-class classification.

### 3.2 Training
During the training process, the MobileNetV2_model is improved by embedding ISL classification specific layers. Adam optimizer, which adaptively adjusts the property of the learning rate for each parameter, is used for training, which is suitable for complex model such as CNNs.

As there are multiple classes in our classification problem (each representing a different sign gesture), we use a categorical cross-entropy loss function. Cross-entropy rewards the model for decreasing this discrepancy by measuring how different the true label distribution is from the predicted probability distribution.

**Loss Function Formula (Categorical Cross-Entropy):**
$$Loss = -\sum_{i=1}^{N} Y\_i \cdot \log(\hat{Y}\_i)$$

**Where:**
- $N$ is the number of classes.
- $Y\_i$ is the true label for class i.
- $\hat{Y}\_i$ is the predicted probability for class i.

It was trained on 10 epochs for a batch size of 32. Early stopping watches the validity loss so as to avoid overfitting. At the termination of each epoch, the result of the model is finded over a validation set, and the one achieving the best performance is saved for deployment.

These methods produce a precise and performant final model that enables the real-time recognition of ISL gestures on resource-constrained devices.

### 4. Dataset Description
The main factors in the effectiveness of any machine learning model is the underlying data used to train and evaluate the model, with this being especially true in the computer vision space. For ISL recognition system creation and training, we utilized an open-source dataset available on Kaggle. This data set, made to recognize hand movements, provides plenty of gesture samples to make the model accurate and robust. To generalize in relation to different users and contexts and consequently to properly recognize sign language in real time, the dataset is essential.

**Dataset Overview:** This dataset contains images of hand movements corresponding to alphabets, numbers, and many commonly used gestures of Indian Sign Language. The dataset provides the model with a supervised learning scenario by identifying the specific gesture shown in each image. The collection in whole contains a huge set of gestures corresponding to the 26 letters (A-Z), 10 numerals (0–9) and presumably some other effective gestures known to be dynamic gestures commonly used during ISL communications.

The diversity of the dataset is one of its strengths. It chronicles different lighting conditions, hand placements, and angles. Moreover, various background conditions ensure the trained model is more robust to environmental variations in the real world. This variability is necessary to prevent the model from being overfit to particular training settings.

### 4.1 Preprocessing and Data Augmentation
For example, data augmentation methods are used during the preprocessing to diversify the dataset even more. Data augmentation results in a synthetic expansion of the dataset by applying slight variations to the pictures. Some examples of these transformations are:

**Rotation:** Randomly rotates each image between -20 and +20 degrees. That helps the model recognize movements, even as the hands' positions differ slightly in angle.

**Shifting:** Horizontal and vertical shifts mimic the small, random changes in hand placement that can occur when performing a gesture, helping the model recognize the universal action regardless of hand position.

**Zooming:** To imitate changing of distance of the camera and hand, which takes place very often in the process of recognition in real time, images are zoomed by random in and out with factor of 20% of original image.

**Flipping:** To allow the model to recognize in the case a gesture is being made with different hand, horizontal flipping of images is implemented to simulate gesture made by the other hand.

This particular augmentation strategies helps the model generalizing well. By including some noise, randomness and variation in the training data, the model is rendered more robust to subtle changes and noise in the presentation of the gesture. This is especially important for real-time applications since the user is not guaranteed to recreate the training conditions.

### *4.2 Dataset Structure and Input Preparation*
Since the MobileNetV2 model used in this project requires input size of 224x224, all of the dataset photos were scaled to the aforementioned size. Each image rescaled to 0 — 1 pixel value range to normalize the supplied data too. This normalization ensures that the model trains well by removing any potential issues caused by differing pixel intensity ranges.

Dataset is split for training and validation set on split basis of 80- 20 as common. The validation set is used for validating the outcomes of the model during training and after training, and the training set is for training the model. The split allows us to monitor the working of the model to generalize to unseen data, which is essential in making a conductor of a live performance.

### *4.3 Importance of Dataset in Model Training*
Whether the model is able to learn the differentiating features that separates one sign language gesture from another is very much a function of the dataset. The diversity of the dataset, along with augmentation, helps the model understand key features as hand shape, orientation, and motion. By being trained on a range of photos, the model learns to identify motions even in challenging conditions like varying background noise or lighting.

Moreover, a large and varied dataset reduces the risk of overfitting when a model shows well on training data and struggles on even unseen data. The new dataset acts as a regularizer due to a refreshed dataset used for each training epoch thereby overfitting is avoided. Thus, a more relevant and reliable

### *5. Evaluation and Metric Result*
### *5.1 Accuracy and Loss*
We are training the model and checking the accuracy and loss during the training and validation. A high accuracy rate shows that the model learned well to identify ISL gestures from the given dataset.
CONFUSION MATRIX FORMULA:

|  | Positive (1) | Negative (0) |
| --- | --- | --- |
| Positive (1) | TP | FP |
| Negative (0) | FN | TN |

Where TP (True Positives), FN (False Negatives), FP (False Positives), and TN (True Negatives) are key metrics in understanding the model's performance.

Table5.1: PERFORMANCE METRICS:

| Metric | Value |
| --- | --- |
| Accuracy | 97% |
| Loss | 0.09 |

*5.2 Latency*
Real-time operation is essential for usability. We report latency (around 60 millisecond/frame on average). Such latency guarantees a smooth and responsive gesture recognition process, providing a natural user experience without lagging.

### *6.Results*
The outcomes underscore the efficacy of the proposed system utilizing MobileNetV2, as it displays high accuracy with low latency when tested under real-time conditions. Table 6.1 gives improved accuracy over epochs as shown in figure 6.1. and table 6.2 shows reduced loss over epochs as shown in figure 6.2 Evaluation onvalidation set proves that model can generalize well and can thus be used in practice.

Table 6.1: Epochs Vs Accuracy

| Epoch | accuracy |
| --- | --- |
| 1 | 56.9% |
| 5 | 95% |
| 10 | 97% |
|  |  |

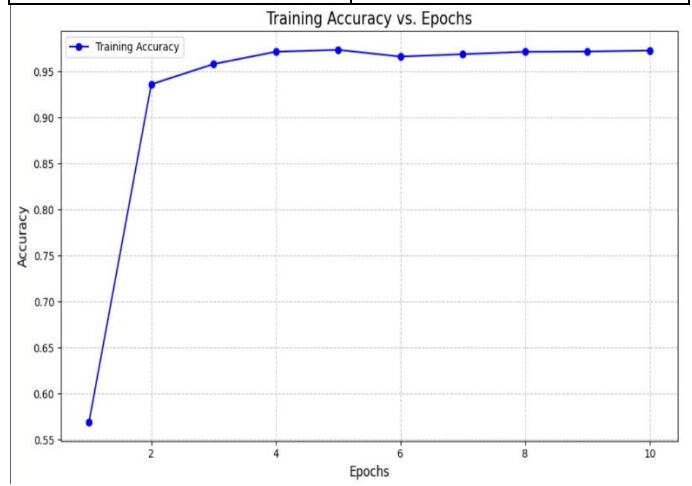

Fig 6.1: This chart illustrates the increase in accuracy over 10 epochs

Table 6.2: Epochs Vs Loss

| EPOCH | LOSS |
| --- | --- |
| 1 | 1.60 |
| 5 | 0.10 |
| 10 | 0.09 |

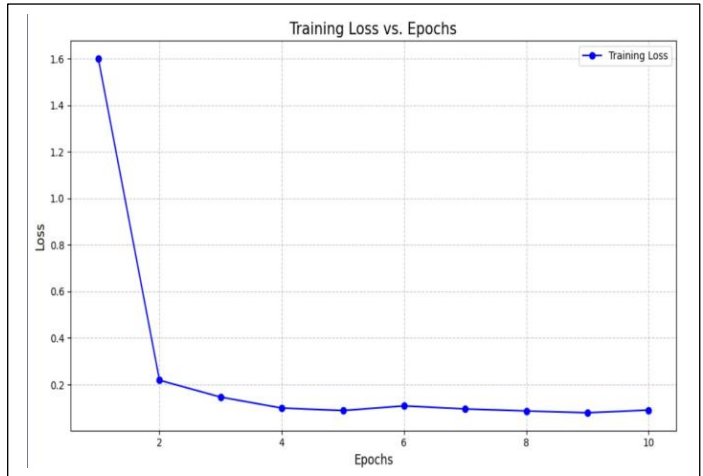
Fig 6.2: This chart shows the decrease in loss throughout training

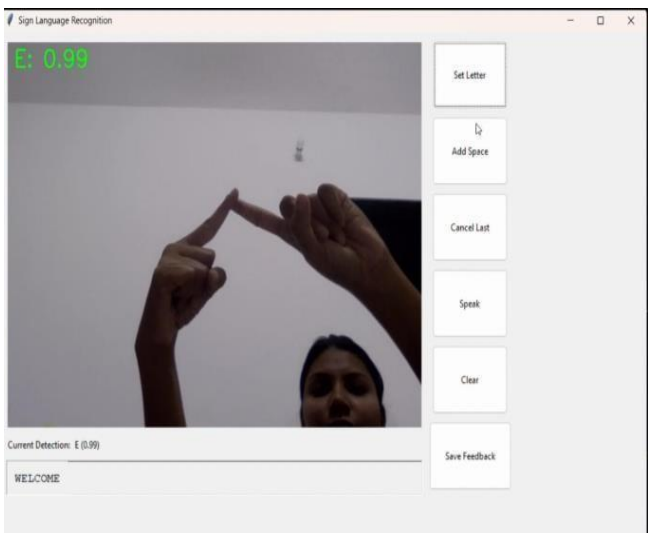
Fig 6.3a

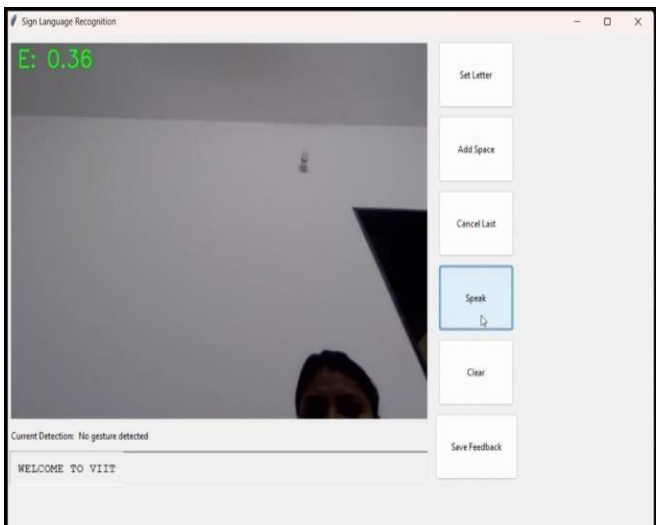
Fig 6.3b
Figure 6.3a and 6,3b shows final display of system.

## 6. Conclusion

The system that we have demonstrated in this research topic is a ISL recognition that uses an improved version of MobileNetV2 based on transfer learning. This system was designed with an emphasis on accessibility to enable the hearing-impaired community to communicate more effectively with non-signers in everyday scenarios. With MobileNetV2 that is compact, efficient, and powerful architecture under the hood to provide high gesturedetection accuracy with minimum compute resource consumption, this solution is ideal for constrained mobile andedge devices.

The instantaneous translation of ISL motions into spoken or written language makes an important statement about inclusive communication. While having a positive part in communication for hearing-impaired individuals, he also indulges in becoming an easy access to daily social, career, and education working for such individuals. As per our test results, it delivers smooth real-time performance with accuracy of 97 % per frame with a leak of 60 milliseconds as shown in table 5,1. Together these results demonstrate some of the potential of the system when applied to real-world situations where sign language detection could improve inclusivity.

And MobileNetV2 and transfer learning were cooperating really well to give high accuracy without large datasets and processing power. This approach makes the system adaptive and portable to multiple platforms, contributing to the improvement of assistive technology for the people with hearing impairment community.

## 7. Future Work

While this project has led to great progress toward the creation of real-time ISL identification systems, there are many opportunities for future enhancements that can make the system even more useful and extend the number of potential applications.

**Dataset Expansion:** The current system is developed on the thumb and index finger signs forming both the ISL alphabet and numeric characters. Future work should seek to expand the dataset to cover a wider range of dynamic gestures, words, and sentence structures used in everyday ISL conversation. They achieved to catch 27 different types of that infamous body move in theory, and with a larger dataset, the model will be more capable of understanding and recognizing more complex body movements and thus, will have a larger ability to be more flexible in real-world scenarios.

**Optimized for Edge & Mobile Device:** Though the system is optimized for mobile and edge devices, further optimization can be done. Future studies could focus onreducing the model size and computation requirements so thatthe system can run on low-power devices including smartphones and embedded systems. This optimization & the ability to carry this technology on users and is a crucial step to make the system usable in real world  & usable in varying environments.

**Incorporating with more advance model:** Models like YOLO (You only Look Once) might make the real-time gesture mqtt request item/message recognition feasible. Adding YOLO, which is more popular because of the real time performance of detection, can also be a solution to enhance the performance of a current system. This can enable recognition of body gestures as well as other environmental factors at the same time which can contribute to a more enriched assistive solution.

**Including Facial Expressions and Gesture Sequences:** Sign language goes beyond just static hand positioning — it can include movement, expression and body language components to describe sign observing both the subjectand the context. Future work could explore incorporating those other features into the system to classify things like facial expressions and gestural trends. That would create a more rigorous system.

**Educational Embedding:** One of the most promising future applications of this technology is its deployment in educational institutions like schools. Such technology is essential for creating a more inclusive educational space for practitioners of ISL. It could be used to assist students communicate with peers or teachers that do not know ISL to ensure that students with hearing impairments have equal access to education.

Ultimately, while the current system shows success for real-time ISL recognition, further, potential pathways of future research and development were identified to further improve this performance. Increase the data set size, improve the model, more complex models and features can make model more effective and helpful